\documentclass[10pt,twocolumn,letterpaper]{article}

 \usepackage{wacv}              

\usepackage{xargs}                      
\usepackage[colorinlistoftodos,prependcaption,textsize=small]{todonotes}
\usepackage{acro}
\usepackage{adjustbox}
\usepackage{threeparttable}
\usepackage{float}
\usepackage{amssymb}  
\usepackage{wasysym}  
\newcommand{\cmark}{\checkmark}
\newcommand{\pmark}{\LEFTcircle} 
\newcommand{\xmark}{$\times$}

\usepackage{multirow}
\usepackage{verbatim}

\acsetup{list/display=used}  
\DeclareAcronym{msb}{
  short = \textsc{MSB}\xspace,
  long  = \textsc{MedSegBenchmarker}\xspace
}
\DeclareAcronym{unet}{
  short = U-Net,
  long  = U-Net
}
\DeclareAcronym{hiformer}{
  short = HiFormer-B,
  long  = HiFormer-B
}
\DeclareAcronym{hiformerRN}{
  short = HiFormer-B$^*$,
  long  = HiFormer-B-RN101
}
\DeclareAcronym{missformer}{
  short = MISSFormer,
  long  = MISSFormer
}
\DeclareAcronym{segformer}{
  short = SegFormer,
  long  = SegFormer-B3
}
\DeclareAcronym{segnext}{
  short = SegNeXt,
  long  = SegNeXt-L
}
\DeclareAcronym{swinumamba}{
  short = SU-Mamba,
  long  = Swin-UMamba
}
\DeclareAcronym{vwmit}{
  short = VW-MiT,
  long  = VWFormer+MiT-B3
}
\DeclareAcronym{vwconv}{
  short = VW-Conv,
  long  = VWFormer+ConvNeXt-S
}
\DeclareAcronym{vwformer}{
  short = VWFormer,
  long  = VWFormer
}
\DeclareAcronym{internimage}{
  short = InternImage,
  long  = InternImage-T
}
\DeclareAcronym{internimageUperhead}{
  short = InternImage,
  long  = InternImage-T+UPerHead
}
\DeclareAcronym{transnext}{
  short = TransNeXt,
  long  = TransNeXt-Tiny
}
\DeclareAcronym{transnextUperhead}{
  short = TransNeXt,
  long  = TransNeXt-Tiny+UPerHead
}
\DeclareAcronym{uperhead}{
  short = UPerHead,
  long  = UPerHead
}
\DeclareAcronym{ukan}{
  short = U-KAN,
  long  = U-KAN-L
}
\DeclareAcronym{cenet}{
  short = CENet$^*$,
  long  = CENet with PVTv2-B3
}
\DeclareAcronym{sota}{
  short = SotA,
  long  = state-of-the-art
}
\DeclareAcronym{ood}{
  short = OOD,
  long  = out-of-distribution
}
\DeclareAcronym{gp}{
  short = GP,
  long  = general-purpose
}
\DeclareAcronym{cnn}{
  short = CNN,
  long  = convolutional neural network
}
\DeclareAcronym{vit}{
  short = ViT,
  long  = vision transformer 
}
\DeclareAcronym{gmac}{
  short = GMAC,
  long  = Giga Multiply-Accumulate Operation
}
\DeclareAcronym{gradcam}{
  short = Grad-CAM,
  long  = Gradient-weighted Class Activation Mapping
}
\DeclareAcronym{hcp}{
  short = HCP,
  long  = healthcare professional
}
\DeclareAcronym{cv}{
  short = CV,
  long  = cross-validation
}
\DeclareAcronym{dl}{
  short = DL,
  long  = deep learning
}
\DeclareAcronym{ml}{
  short = ML,
  long  = machine learning
}
\DeclareAcronym{gt}{
  short = GT,
  long  = ground truth
}
\DeclareAcronym{bkai}{
  short = NeoPolyp,
  long  = BKAI-IGH NeoPolyp Small
}
\DeclareAcronym{isic}{
  short = ISIC'18,
  long  = ISIC'18
}
\DeclareAcronym{camus}{
  short = CAMUS,
  long  = CAMUS 
}
\DeclareAcronym{mis}{
  short = MIS,
  long  = medical image segmentation
}
\DeclareAcronym{ss}{
  short = SS,
  long  = semantic segmentation
}
\DeclareAcronym{xai}{
  short = XAI,
  long  = explainability
}
\DeclareAcronym{gpvm}{
  short = GP-VM,
  long  = general-purpose vision model
}
\DeclareAcronym{sma}{
  short = SMA,
  long  = specialized medical segmentation architecture
}
\DeclareAcronym{kan}{
  short = KAN,
  long  = Kolmogorov–Arnold Network
}
\DeclareAcronym{ci}{
  short = CI,
  long  = confidence interval
}
\DeclareAcronym{rng}{
  short = RNG,
  long  = random-number-generator
}
\DeclareAcronym{hpo}{
  short = HPO,
  long  = hyperparameter optimization
}

\newif\ifanonymous
\anonymoustrue        
\ifanonymous
  \newcommand{\githuburl}{https://anonymous.4open.science/r/MedSegBenchmarker-6D4C/}
\else
  \newcommand{\githuburl}{https://github.com/VanessaBorst/MedSegBenchmarker/}
\fi

\newcommand{\toolname}{\textsc{MedSegBenchmarker}\xspace}
\newcommand{\toolnameShort}{\textsc{MSB}\xspace}

\definecolor{wacvblue}{rgb}{0.21,0.49,0.74}
\usepackage[pagebackref,breaklinks,colorlinks,allcolors=wacvblue]{hyperref}

\def\wacvPaperID{90} 
\def\confName{WACV}
\def\confYear{2027}

\title{MedSegBenchmarker: A Raw-Count-First Framework for\\ Controlled 2D Medical Image Segmentation Benchmarks}

\author{
\begin{tabular}{ccccc}
Vanessa Borst &
Lukas Horn &
Daniel Grillmeyer&
Thomas Prantl&
Samuel Kounev
\end{tabular}
\\[8pt]
University of Würzburg, Germany \\
{\tt\small vanessa.borst@uni-wuerzburg.de}
}

\begin{document}
\maketitle
\begin{abstract}
Despite rapid advances in \ac{mis}, fair and reproducible comparisons of segmentation models remain challenging due to heterogeneous datasets, inconsistent evaluation protocols, and rapidly evolving architectures. In particular, comparisons often implicitly assume that model rankings are invariant to data partitioning, preprocessing, metric aggregation, uncertainty estimation, and computational constraints. The lack of extensible and unified evaluation frameworks further limits systematic investigation of new models, datasets, and training paradigms.
We present \toolname (\toolnameShort), a configuration-driven framework for controlled benchmarking of 2D \ac{mis}. It integrates duplicate and near-duplicate detection during data preparation, group-aware data splitting, 
layered YAML study specifications, resumable training, \ac{hpo}, \ac{cv}, and checkpoint-based final evaluation. Rather than retaining only aggregate performance measures, \toolnameShort exports sample- and class-level pixel counts and predictions together with the associated evaluation context. These elementary artifacts enable post-hoc analyses 
without requiring repeated inference.
We demonstrate \toolnameShort in a case study involving three heterogeneous 2D datasets and multiple \ac{mis} and \acl{gp} vision models evaluated at $256$- and $512$-pixel input resolutions. Reaggregation of identical predictions changes the top-ranked architecture in three of six dataset--resolution settings, despite high rank correlations between aggregation strategies. Increasing input resolution produces model- and dataset-dependent performance gains and losses that must be considered alongside empirically measured inference complexity. 
These results show that seemingly minor choices in evaluation and experimental setup can affect benchmark conclusions. \toolnameShort, available at \href{\githuburl}{GitHub}, provides a practical and extensible basis for making benchmark conditions and evaluation choices explicit and reproducible.
\end{abstract}

\section{Introduction}
\label{sec:intro}

In \ac{mis}, progress is commonly communicated through improved metric scores attributed to a new encoder~\cite{chen2024TransUNet, hu2023SwinUnet, jiacheng2025VM-UNet, tang2022SwinTransformers}, decoder~\cite{chen2024TransUNet, hu2023SwinUnet, mostafijur2023CascadedAttentionDecoding, mostafijur2024EMCAD, jiacheng2025VM-UNet}, attention mechanism~\cite{mostafijur2023CascadedAttentionDecoding, mostafijur2024EMCAD}, or pretraining strategy~\cite{zhou2021ModelsGenesis, tang2022SwinTransformers, ma2024SegmentAnythingMedical}. That score, however, is inseparable from the experiment that produced it. Dataset curation, patient separation, input resolution, augmentation, optimization budget, checkpoint selection, and metric aggregation can all affect the reported result. Empirical work has documented substantial variability in deep-learning segmentation pipelines~\cite{renard2020variability}, while broader analyses identify experimental design and reporting deficiencies as recurring threats to reliable medical-imaging research~\cite{varoquaux2022machine,pineau2021improving}. In a review of recent \ac{mis} papers, more than half reported no measure of variability and fewer than one percent reported \acp{ci}~\cite{christodoulou2024confidence}.

Evaluation choices are particularly consequential when target structures are rare or heterogeneous. 
Pooling pixels across a test set gives greater weight to large structures, whereas averaging image-level scores weights each image equally; class-macro aggregation instead favors rare labels. These estimators answer different questions rather than providing interchangeable versions of one result. Likewise, higher image resolution may better represent small details while imposing model-dependent compute and memory costs. 
Biomedical challenge rankings can be unstable under changes in test cases and aggregation schemes~\cite{maierhein2018rankings}; recent guidance emphasizes matching metrics to the scientific question and properties of the target structures~\cite{reinke2024metrics}.

These issues reflect a broader reproducibility crisis affecting scientific research across disciplines~\cite{baker1500ScientistsLift2016, Stupple2019reproducibility}. Kapoor~and~Narayanan~\cite{kapoor2023leakage} documented reproducibility failures and methodological pitfalls, including data leakage, across 648 papers spanning medicine, neuroimaging, radiology, and dermatology, contributing in some cases to substantially overoptimistic performance claims. Such discrepancies can impede scientific progress, waste computational and human resources, and increase the risk of deploying unreliable systems in high-stakes settings.

Concerns about the validity of \ac{ml} evaluations are neither new nor confined to a single pipeline stage. Semmelrock~et~al.~\cite{semmelrock2025reproducibility} categorize reproducibility barriers into description, code, data, and experiment, concluding that the reliability of \ac{ml} evaluation remains underemphasized and lacks established best practices, particularly when \ac{ml} frameworks are used by domain experts rather than data scientists. 
Liao~et~al.~\cite{liaoAreWeLearning2021} show in a meta-review that pitfalls arise throughout the \ac{ml} lifecycle, including model training, baseline selection, and dataset and metric choice, indicating that no single stage can be addressed in isolation. Hutchinson~et~al.~\cite{hutchinsonEvaluationGapsMachine2022} further confirm this in an analysis of 200 papers from top-tier NLP and computer vision venues, finding a systematic gap between what should be evaluated and what is actually evaluated.

Dataset collections, benchmarks, and software frameworks address important parts of this landscape. The Medical Segmentation Decathlon~\cite{antonelli2022msd} and MedSegBench~\cite{kus2024medsegbench} provide curated evaluation tasks; MONAI~\cite{cardoso2022monai} and TorchIO\cite{perezgarcia2021torchio} provide reusable medical-imaging components; and nnU-Net automatically configures a strong segmentation method for a new task~\cite{isensee2021nnunet}. Our scope is complementary: we focus on controlled comparisons among heterogeneous segmentation architectures and on retaining the elementary evidence needed to examine how evaluation choices affect their conclusions.

We present \toolname, a configuration-driven research framework for controlled 2D \ac{mis} benchmarks. 
The current implementation targets modalities acquired as 2D images, including dermoscopy, endoscopy, and ultrasound. It supports duplicate and near-duplicate diagnostics during data preparation, as well as group-aware splitting when group metadata are supplied. Layered YAML specifications describe shared, dataset-specific, and architecture-specific settings; execution covers single runs, declared Cartesian grid or Hyperband searches, and $K$-fold \ac{cv}, with resumable run state. Final evaluation reloads the selected checkpoint and exports foreground TP, FP, FN, and TN counts for each sample--class pair, together with the relevant split, fold, run, evaluation, and prediction-semantics fields. These counts are then the basis for subsequent aggregation, uncertainty estimation, figures, and \LaTeX{} tables.

To characterize the conditions of a model comparison, we introduce the \emph{benchmark contract} as a conceptual representation of factors that may influence evaluation outcomes: dataset and labels, split policy, preprocessing, data augmentation, model settings, optimization and checkpoint-selection policy, inference rules, and evaluation procedure. The contract does not certify that a comparison is fair; rather, it provides a structured means to record and inspect shared controls, justified differences, and architecture-specific constraints across the configurations and artifacts used in a study.

Overall, our contributions can be summarized as follows:

\begin{itemize}
    \item \textbf{Configuration-driven benchmarking tool:} 
    We design a modular workflow connecting dataset diagnostics and splitting, layered study specifications, resumable execution, checkpoint-backed final evaluation, and post-hoc analysis for 2D \ac{mis}. \toolname produces reproducible artifacts, including raw \ac{cv} results and scripts for regenerating analyses and reports.

    \item \textbf{Raw-count-first evaluation and benchmark contract:} 
    Sample--class confusion counts retain the information required for several count-based aggregation views and paired resampling analyses after inference, while the benchmark contract organizes the methodological choices needed to interpret a comparison.
    By mapping the contract directly to the implementation, \toolnameShort enables systematic and reproducible comparisons of heterogeneous architectures for \ac{mis}.
    
    \item \textbf{Case study of evaluation sensitivity:} 
    We demonstrate \toolnameShort in a case study covering eleven architectures and three heterogeneous 2D \ac{mis} datasets at two input resolutions of $256\times256$ and $512\times512$. The study reaggregates fixed raw counts under global-micro, image-macro, and class-macro Dice, compares paired resolution changes, and reports model-only inference complexity. Its purpose is to illustrate the consequences of evaluation design, rather than to establish an universal architecture ranking.

\end{itemize}

\section{Related Work}
\label{sec:related_work}

Existing software addresses complementary parts of a segmentation study: reusable medical-imaging components, self-configuring task-specific pipelines, benchmark datasets, and general model implementations. Table~\ref{tab:framework_comparison} retains a scope-oriented comparison of representative systems, with Tier~1 comprising medical image frameworks, Tier~2 consisting of benchmarking studies or datasets, and the final tier (G) encompassing general-purpose semantic segmentation toolkits.
\begin{table*}[htbp]
    \centering
    \caption{Comparison of segmentation frameworks and benchmarks (\cmark = fulfilled, \pmark = partially fulfilled, \xmark = not fulfilled).}
    \label{tab:framework_comparison}
    \resizebox{\textwidth}{!}{%
    \begin{tabular}{c l c c c c c c c c c c}
        \toprule
        Tier & Framework / Benchmark & Counts & Ext. & Config. & \ac{hpo} & Audit & OOD & XAI & Profil. & Plotting & Notifs. \\
        \midrule
        --- & \textbf{\toolname (\toolnameShort)} & \cmark & \cmark & \cmark & \cmark & \cmark & \cmark & \cmark & \cmark & \cmark & \cmark \\
        \midrule
        \multirow{4}{*}{1} & nnU-Net~\cite{isensee2021nnunet}                           & \cmark & \cmark & \cmark & \cmark & \xmark & \xmark & \xmark & \pmark & \pmark & \xmark \\ 
                           & MONAI~\cite{cardoso2022monai}                              & \pmark & \cmark & \cmark & \cmark & \xmark & \xmark & \cmark & \xmark & \cmark & \pmark \\ 
                           & TorchIO~\cite{perezgarcia2021torchio}                      & \xmark & \cmark & \pmark & \xmark & \xmark & \xmark & \xmark & \xmark & \xmark & \xmark \\ 
                           & TorchXrayVision~\cite{Cohen2022xrv}                        & \xmark & \cmark & \xmark & \xmark & \pmark & \cmark & \xmark & \xmark & \xmark & \xmark \\
        \midrule
        \multirow{4}{*}{2} & MedSegBench~\cite{kus2024medsegbench}                      & \xmark & \cmark & \pmark & \xmark & \xmark & \xmark & \xmark & \xmark & \xmark & \xmark \\
                           & Medical Segmentation Decathlon~\cite{antonelli2022msd}     & \xmark & \xmark & \xmark & \xmark & \xmark & \cmark & \xmark & \xmark & \xmark & \xmark \\
                           & TotalSegmentator~\cite{Wasserthal_2023}                    & \xmark & \cmark & \pmark & \cmark & \xmark & \xmark & \xmark & \xmark & \pmark & \xmark \\ 
                           & FeTS Challenge Toolkit~\cite{zenk_towards_2025,fets_tool}  & \xmark & \cmark & \cmark & \xmark & - & \cmark & \xmark & \xmark & \xmark & \xmark \\ 
        \midrule
        \multirow{4}{*}{G} & PaddleSeg~\cite{liu2021paddleseg}                              & \xmark & \cmark & \cmark & \cmark & \xmark & \xmark & \xmark & \cmark & \cmark & \xmark \\ 
                           & Segmentation Models PyTorch~\cite{Iakubovskii:2019}            & \pmark & \cmark & \cmark & \xmark & \xmark & \xmark & \xmark & \xmark & \xmark & \xmark \\ 
                           & MMSegmentation~\cite{mmseg2020}                                & \xmark & \cmark & \cmark & \xmark & \xmark & \xmark & \cmark & \cmark & \cmark & \xmark \\ 
                           & Transformers Hugging Face~\cite{wolf-etal-2020-transformers}   & \xmark & \cmark & \cmark & \cmark & \pmark & \pmark & \xmark & \xmark & \xmark & \pmark \\ 
        \bottomrule
    \end{tabular}%
    }
\end{table*}
In particular, we consider the following categories:
\begin{enumerate*}[label=(\Roman*)]
\item Counts: Persistently exported sample-level confusion-count evidence. 
\item Extensibility: Easy inclusion of new models, datasets, and metrics.
\item Configurability: Fine-grained control over architectures, augmentations, optimizers, etc.
\item \ac{hpo}: Automatic hyperparameter tuning (e.g., grid search).
\item Data Audit: Duplicate or leakage diagnostics; not necessarily automatic prevention.
\item \Ac{ood} Analysis: workflow for evaluation on an external test dataset. 
\item \Ac{xai}: Integrated qualitative XAI outputs, such as Grad-CAM~\cite{gradcam}.
\item Profiling: Integrated model-complexity, latency, or memory profiling.
\item Plotting: Utilities and capabilities for generating plots etc. 
\item Notification: Push-based status notifications upon benchmark completion.
\end{enumerate*}
All listed systems are Python-based and can, in principle, be combined with specialized third-party frameworks, for example for \ac{hpo}. Such integrations require users to select, connect, and maintain an additional tool within the experimental workflow and are therefore denoted by ``\xmark''. A ``\cmark'' is reserved for functionality implemented within the system's own workflow and exposed through interfaces; study-specific configuration may still be required.

\textbf{Medical Image Segmentation Frameworks.}
Prominent \ac{mis} frameworks include nnU-Net~\cite{isensee2021nnunet}, MONAI~\cite{cardoso2022monai}, TorchIO~\cite{perezgarcia2021torchio} and TorchXrayVision~\cite{Cohen2022xrv}. 
MONAI provides composable networks, transforms, losses, and inference utilities for healthcare imaging~\cite{cardoso2022monai}, while TorchIO focuses on image loading, preprocessing, augmentation, and patch sampling~\cite{perezgarcia2021torchio}. nnU-Net addresses a different, adjacent objective: it derives a strong task-specific pipeline from dataset characteristics~\cite{isensee2021nnunet}. TorchXRayVision provides a standardized interface to chest X-ray datasets, preprocessing routines, and pretrained deep-learning models~\cite{Cohen2022xrv}.
These systems provide valuable primitives, reference implementations, and task-specific automation. \toolnameShort operates at a different layer: it coordinates a declared comparison across heterogeneous model adapters, records the checkpoint used for final evaluation, and retains sample--class count artifacts for subsequent analysis. It is therefore complementary to, rather than a replacement for, their model, transform, and training components.

MONAI is the closest comparator in scope, combining broad medical-imaging primitives, visualization, and extensibility. Nevertheless, it is a component framework rather than a benchmark-control layer: its bundles may contain executable Python expressions, while external-dataset workflows and the use of optimization libraries such as Optuna~\cite{akiba2019optuna} or Ray Tune~\cite{liaw2018tune} require study-specific integration. \toolnameShort instead exposes seed control and an optional deterministic mode through explicit YAML keys (\texttt{random\_seed} and \texttt{full\_determinism}). Moreover, it provides declared grid/Hyperband execution, checkpoint-backed raw-count export, duplicate diagnostics, and separate efficiency profiling. These facilities make relevant choices more explicit and inspectable.

\textbf{Benchmarking Studies and Datasets.}
The Medical Segmentation Decathlon established a collection of ten 3D tasks for evaluating algorithmic generalization~\cite{antonelli2022msd}. MedSegBench broadens task and modality coverage through a curated collection of 2D and 3D \ac{mis} datasets~\cite{kus2024medsegbench}. TotalSegmentator provides a widely used task-specific segmentation application~\cite{Wasserthal_2023}, while the FeTS Toolkit supports challenge-oriented federated-learning workflows~\cite{zenk_towards_2025,fets_tool}. Such resources define valuable tasks, data, and reference settings. However, their primary focus is not on providing general-purpose benchmarking capabilities, such as automated \ac{hpo}, standardized visualization, or comprehensive performance evaluation across heterogeneous models. Consequently, such functionality may require additional implementation or integration within the respective workflow.
In contrast, \toolnameShort provides dataset-agnostic infrastructure for configuring and inspecting comparative studies once a dataset has been converted to its indexed image--mask schema. This design allows existing benchmark datasets to be integrated without requiring the benchmarking framework itself to be tailored to a specific dataset or task.

\textbf{General Segmentation Frameworks.}
General-purpose segmentation frameworks like PaddleSeg~\cite{liu2021paddleseg}, Segmentation Models PyTorch~\cite{Iakubovskii:2019}, MMSegmentation~\cite{mmseg2020}, and Hugging Face Transformers~\cite{wolf-etal-2020-transformers} offer broad model zoos and strong community support. 
\toolnameShort is not intended to replace these libraries or to provide a general model zoo. Instead, it supplies study-level control for declared 2D \ac{mis} comparisons, while architectures remain available through model adapters and can be implemented using general-purpose framework components.
\section{MedSegBenchmarker Toolbox}
\label{sec:toolbox}

\subsection{Design Rationale}

\Ac{mis} studies commonly combine dataset preparation, split generation, model training, \ac{hpo}, \ac{cv}, post-hoc analysis, and figure generation through study-specific scripts. Such fragmentation can obscure the relationship between a reported result and the experimental choices that produced it, including the data split, selected checkpoint, decision threshold, and aggregation rule. It can therefore impede the inspection, reproduction, and comparison of results.

In contrast, \toolname was developed as a Python/PyTorch toolbox for controlled end-to-end benchmarks in supervised 2D \ac{mis}. It supports binary and multi-class tasks from an indexed image--mask dataset through training and final evaluation to reusable, sample-level evaluation data. 
Overall, its design is guided by five principles: 
\textbf{(I) Transparency}, through human-readable YAML configurations, logging, and structured outputs; 
\textbf{(II) Reproducibility} support, through seed control, checkpointed states, and an optional deterministic execution mode; 
\textbf{(III) Modularity}, through separate data, training, evaluation, and analysis components; 
\textbf{(IV) Extensibility}, through common dataset and model interfaces; and 
\textbf{(V) Traceability}, by linking final-evaluation counts to their configuration hash, checkpoint reference, and prediction semantics. 

Together, these principles support configurable benchmarking across \ac{dl} architectures. Configurations declare architectures, transforms, optimizers, loss functions, checkpoint policies, and evaluation settings. The shared evaluator uses common metric definitions and records binary decision thresholds or multi-class argmax semantics with the final counts. When experiments are run under a stated common protocol, these facilities make the resulting comparison more transparent and inspectable; they do not by themselves establish that the protocol is fair.

\subsection{Benchmark Contract}
\label{sec:benchmark_contract}

We use the term \emph{benchmark contract} for the methodological choices that define the conditions under which performance is estimated and compared:
\begin{equation}
\mathcal{C} =
(\mathcal{D}, \mathcal{S}, \mathcal{P}, \mathcal{M},
\mathcal{O}, \mathcal{I}, \mathcal{E}).
\label{eq:benchmark_contract}
\end{equation}
Here, $\mathcal{D}$ denotes the dataset and label mapping; $\mathcal{S}$ the split and, where available, grouping policy; $\mathcal{P}$ preprocessing and augmentation; and $\mathcal{M}$ the model architecture and its declared settings. $\mathcal{O}$ comprises optimization and checkpoint selection, $\mathcal{I}$ the output interpretation and decision rule, and $\mathcal{E}$ the retained elementary outputs, metric definitions, aggregation, and uncertainty procedure.

The contract is a conceptual representation, not a universal manifest format. In \toolnameShort, its elements are declared across YAML configurations, the stored $K$-fold state, checkpoint records, and final-evaluation artifacts. The configuration generator combines global, architecture-specific, and dataset-specific settings into runnable YAML files, making experimental choices such as normalization, batch size, or precision settings visible rather than hidden in orchestration code. The raw-count artifact then binds an evaluation to a configuration hash, an explicit checkpoint reference, and a decision rule. Thus, the toolbox makes many comparison choices inspectable, but it 
does not certify that all choices in $\mathcal{C}$ are scientifically appropriate. Those choices should be fixed or justified before inspecting final results.

\subsection{Five-Stage Workflow}

\begin{figure*}[t]
  \centering
  \includegraphics[width=\linewidth]{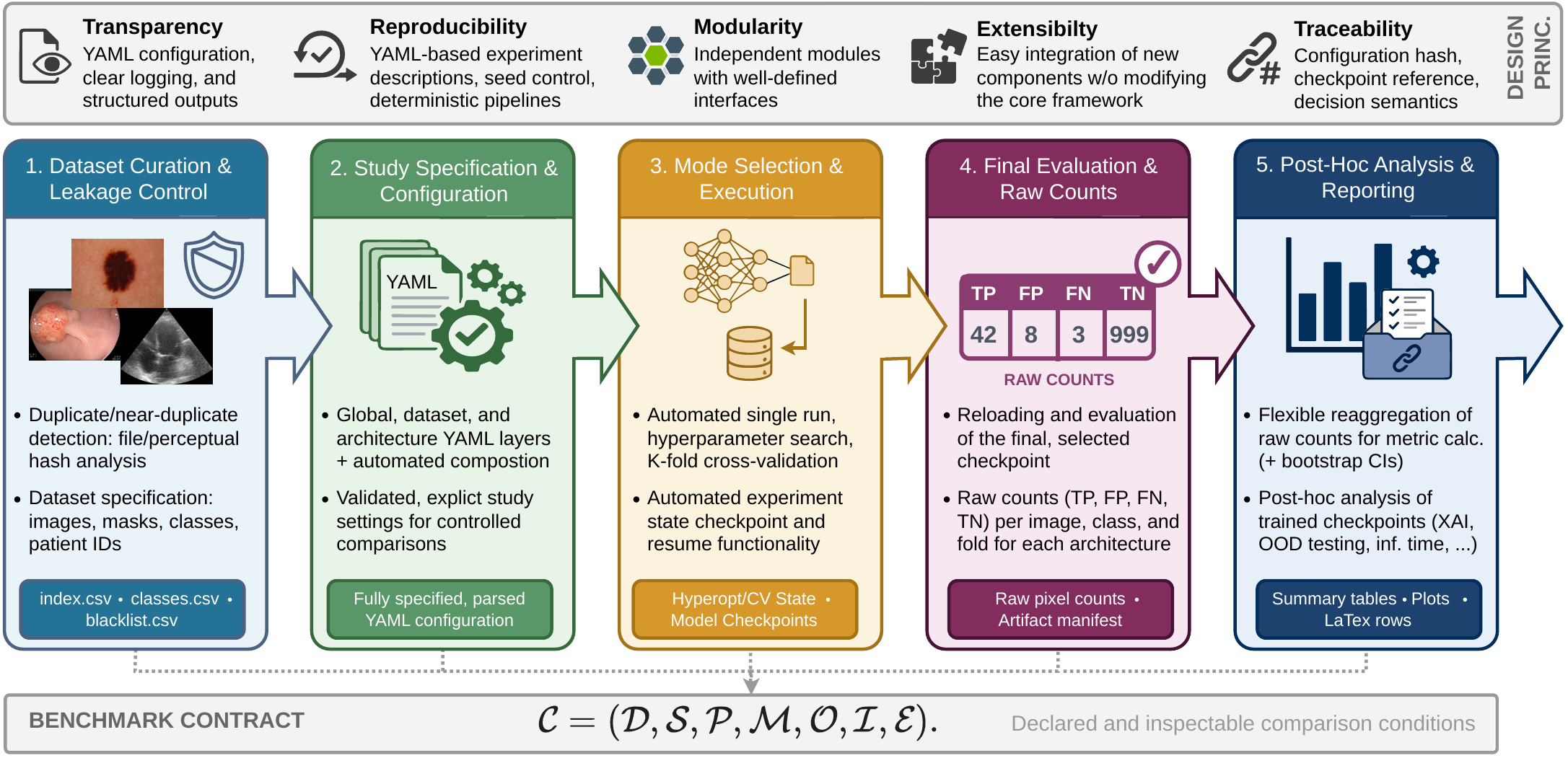}
  \caption{Five-stage study workflow in \toolname. 
  }
  \label{fig:medseg_benchmarker_pipeline}
\end{figure*}

Rather than treating training as an isolated task, \toolname organizes a study around persisted configurations, states, checkpoints, and evaluation outputs. Fig.~\ref{fig:medseg_benchmarker_pipeline} summarizes the stages: \raisebox{.5pt}{\textcircled{\raisebox{-.9pt}{1}}} dataset preparation and audit; \raisebox{.5pt}{\textcircled{\raisebox{-.9pt}{2}}} YAML-based study specification; \raisebox{.5pt}{\textcircled{\raisebox{-.9pt}{3}}} training, \ac{hpo}, or $K$-fold \ac{cv}; \raisebox{.5pt}{\textcircled{\raisebox{-.9pt}{4}}} checkpoint-backed final evaluation with raw-count export; and \raisebox{.5pt}{\textcircled{\raisebox{-.9pt}{5}}} post-hoc analysis and reporting.

\textbf{1. Dataset Curation and Leakage Control.}
The common dataset interface uses \texttt{index.csv} to associate image and mask files with their split, dimensions, and file types, and \texttt{classes.csv} to map source mask values to the training classes. Dataset-specific converters are available for some public datasets as an example, while any custom dataset following this layout can use the same training path. Fixed splits may be read from the index, and seeded custom splits are supported. In $K$-fold \ac{cv}, a configured group column keeps all samples of a known group, such as a patient, in the same fold.
\acs{msb} also provides optional duplicate-assessment utilities. They report byte-identical files using SHA-256, identical perceptual hashes, and candidate near-duplicate groups based on perceptual-hash distance; an Excel report and a graphical review tool support manual inspection of the candidate images. A mask-drawing utility further supports visual dataset inspection. These utilities identify potential leakage routes but do not prove patient identity or automatically remove samples: review decisions must be applied during curation, and unknown cross-subject dependencies remain a dataset-specific risk.

\textbf{2. Study Specification and Configuration.}
YAML specifications declare the dataset, transforms, architecture, optimizer, loss, learning-rate (LR) schedule, checkpoint policy, and evaluation settings. 
The configuration generator can compose global controls with architecture- and dataset-specific layers, yielding an explicit configuration for each experiment. This provides a practical place to state legitimate model accommodations, while retaining the common settings that define a comparison.

\textbf{3. Controlled and Resumable Execution.}
A common \texttt{Segmentor} interface separates orchestration from model implementation. The current registry covers convolutional, transformer, hybrid, and state-space-inspired designs (cf. Section~\ref{sec:case_study}). 
Some model adapters require optional packages or native extensions, so registration alone does not claim support on every platform. The execution layer supports single runs, Cartesian grid search, Hyperband, and $K$-fold \ac{cv}; the latter persists generated fold indices. Training checkpoints retain model, optimizer, scheduler, gradient-scaler, early-stopping, metric-tracking, and random-number-generator states for resumption. Deterministic execution can be configured, although exact numerical equality may still depend on hardware and library versions.

\textbf{4. Checkpoint-backed Raw-Count Evaluation.}
Final evaluation explicitly reloads the configured checkpoint role (best validation, latest, or both). Its checkpoint record stores the epoch, selection split, metric, value, and a SHA-256 content hash. By default, final evaluation writes one raw-count record for each image--class pair. In addition to true positives ($TP$), false positives ($FP$), false negatives ($FN$), and true negatives ($TN$), the record contains intersection, reference-mask, predicted-mask, and union counts; dataset, split, fold, run, model, checkpoint, image, and class identifiers; and the prediction semantics. Binary records retain the decision threshold and whether the model output was interpreted as logits or probabilities; multi-class records retain the argmax rule. 
These counts are sufficient to recompute, for example, DSC, which is defined as  \(2TP/(2TP+FP+FN)\), or IoU as \(TP/(TP+FP+FN)\).
Crucially, the counts can be pooled before evaluating the respective formula. They therefore preserve the information required for foreground-pixel pooling as well as image-weighted, class-weighted, or fold-wise summaries. 

For each fold of a $K$-fold evaluation, a compressed PyTorch artifact  
additionally retains the reference masks and, depending on the configured storage option, raw model outputs (logits), hard predictions, or both. The artifact links image filenames, the checkpoint path, the class mapping, and the declared prediction semantics, including the decision threshold where applicable, to the stored artifact. It therefore preserves the per-fold segmentations needed to compute additional geometry- or contour-based measures later without repeating inference. When logits are retained, hard predictions can also be regenerated under an alternative documented decision threshold; this enables subsequent threshold or calibration analyses.

\textbf{5. Post-hoc Aggregation and Reporting.}
The post-hoc layer distinguishes analyses of retained artifacts from procedures that require model access. Without reloading checkpoints or repeating inference, raw-count artifacts can be validated and consolidated with duplicate-row detection, then re-aggregated to derive per-image--class, per-sample foreground, pooled-class, and pooled-foreground results. The provided analysis recomputes count-based metrics, including Dice, IoU, precision, and recall, estimates image-level bootstrap intervals, and generates PNG/SVG figures and \LaTeX{} table entries. When the optional fold-wise output artifacts are retained, additional output-based measures and visualizations can likewise be computed post hoc.
Checkpoint-backed utilities are required only for analyses that need a model forward pass. They support Grad-CAM-style maps as diagnostic output, model-only complexity and inference profiling, including latency and peak GPU-memory measurements after warm-up, and evaluation of $K$-fold checkpoints on a declared external test dataset.

\section{Case Study}
\label{sec:case_study}

This case study illustrates the analytical value of retaining evaluation artifacts after model training has finished, using a fixed collection of checkpoints and sample-level pixel counts to ask whether benchmark conclusions depend on aggregation, input resolution, computational constraints, or transfer to an external dataset (\ac{ood} generalization).

\subsection{Datasets, Model Cohort, and Artifact Scope}
\label{ssec:case_study_extent}

We use three heterogeneous 2D \ac{mis} datasets: \acs{isic} dermoscopic lesion~\cite{codella2019isic,tschandl2018ham10000}, \acl{bkai} endoscopic polyp~\cite{bkaiigh2021}, and \acs{camus} echocardiographic segmentation~\cite{leclerc2019camus}. After duplicate assessment and removal, the datasets contain 3,565, 945, and 1,996 images, respectively. \acs{isic} is binary; \ac{bkai} comprises background, neoplastic polyp, and non-neoplastic polyp; and \acs{camus} comprises background, left-ventricular cavity, myocardium, and left atrium. \acs{camus} uses patient-level folds, whereas \acs{isic} and \ac{bkai} use image-level folds. 
Building on the selection rationale of prior work~\cite{Borst2026bmvc}, the benchmarked model cohort comprises \mbox{\acs{unet}}~\cite{ronneberger2015unet}, \mbox{\acs{hiformerRN}}~\cite{heidari2023hiformer}, \mbox{\acs{missformer}}~\cite{huang2022missformer}, \ac{swinumamba}~\cite{liu2024swinumamba}, \mbox{\acs{cenet}}~\cite{bozorgpour2025cenet}, \ac{segformer}~\cite{xie2021segformer}, \mbox{\acl{segnext}} (\acs{segnext})~\cite{guo2022segnext}, \mbox{\acs{vwformer}} with MiT-B3 backbone (\acs{vwmit}), \mbox{\acs{vwformer}} with ConvNeXt-S backbone (\acs{vwconv}), \mbox{\ac{internimage}}~\cite{wang2023internimage} with UPerHead~\cite{xiao2018upernet}, and \mbox{\ac{transnext}}~\cite{shi2024transnext} with UPerHead. 

For this case study, we reused the $256\times256$ final-evaluation artifacts generated as part of prior work~\cite{Borst2026bmvc} and reproduced the same experimental protocol at $512\times512$, including hyperparameter selection followed by five-fold \ac{cv}. This protocol was applied to all methods except \ac{cenet}, which exceeded the available GPU memory at the increased input resolution. Consequently, matched five-fold checkpoints and final-evaluation artifacts are available at both resolutions for ten architectures, while \ac{cenet} was excluded from analyses requiring matched-resolution results.
Details of the architectures and setup are provided in the Appendix.

\subsection{Analysis Protocol}
\label{ssec:case_study_protocol}

All post-hoc analyses described below reuse the saved raw counts and model checkpoints and do not involve retraining any method. The following provides a brief summary of the conducted analyses, while detailed descriptions and additional results are provided in the supplementary material.

\textbf{Aggregation Sensitivity.}
We recomputed foreground Dice under three defensible aggregation rules: \emph{global micro}, which pools foreground counts before calculating Dice; \emph{image macro}, which first pools foreground classes per image and then averages image-level Dice; and \emph{class macro}, which averages Dice after pooling counts within each foreground class. Images without foreground in both prediction and reference have an undefined Dice denominator and are therefore excluded from image-macro Dice. Architectures were ranked separately for every dataset, resolution, and aggregation rule. We report rank correlation, changed ranks, and whether the leading architecture changes between rules.

\textbf{Input-Resolution Sensitivity.}
We compared $256\times256$ and $512\times512$ predictions matched by dataset, architecture, CV fold, and test image. The primary estimand was the paired global-micro change \mbox{$\Delta\mathrm{DSC}=\mathrm{DSC}_{512}-\mathrm{DSC}_{256}$}, with 95\% percentile \acp{ci} estimated from 1,000 paired, fold-stratified bootstrap replicates (excluding \ac{cenet}).

\textbf{Inference Efficiency.}
We measured device-resident, model-only forward-pass efficiency for batch size one on an NVIDIA A100 MIG 40\,GB partition. Timing used the first \ac{bkai} \ac{cv} checkpoint per architecture and resolution, 100 warm-up passes, 1,000 timed passes, and five repetitions. The timed region excludes preprocessing, transfer, decoding, postprocessing, and I/O. We report parameter count, GMACs, median latency, derived throughput, and maximum allocated CUDA memory.

\textbf{External Same-task Cross-Dataset Evaluation.}
For same-task external transfer, we evaluated all five \ac{bkai} \ac{cv} checkpoints for each available architecture--resolution combination on the 1,000-image, test-only Kvasir-SEG dataset~\cite{jha2019kvasir}, without retraining, fine-tuning, target-domain threshold selection, or target-informed model selection. Both source foreground classes were merged into a binary \emph{polyp} label before fold-wise scoring and ensembling. We then formed a pixel-wise strict-majority ensemble, assigning foreground when at least three of five folds predicted polyp, and recomputed foreground Dice, IoU, precision, and recall from retained per-image pixel counts for individual folds and ensembles. The accompanying $256\times256$ qualitative example shows the ensemble prediction and a fold-aggregated \acs{gradcam} diagnostic; details of the ontology mapping, attribution aggregation, and representative-image selection are provided in the Appendix.

\begin{figure*}[ht]
  \centering
  \includegraphics[width=\textwidth]{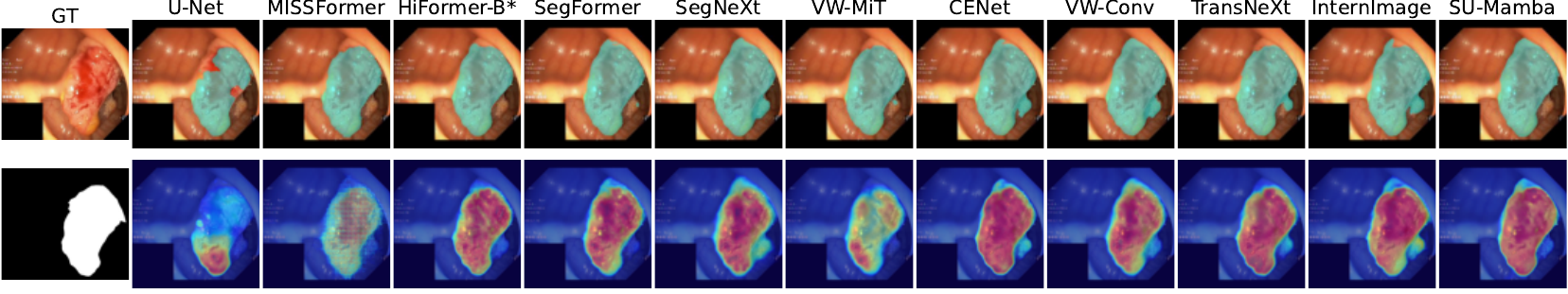}
  \caption{Representative $256\times256$ Kvasir-SEG example with ensemble masks and fold-aggregated Grad-CAM maps for each architecture.}
  \label{fig:case_study:kvasir_ood:XAI}
\end{figure*}
 
\subsection{Case Study Results}
\label{ssec:case_study_results}

\textbf{Reaggregation can change rankings.}
Table~\ref{tab:aggregation-sensitivity} compares the three aggregation rules pairwise. Despite high rank correlations ($\rho=0.93$--$1.00$), reaggregation changes up to seven architecture ranks per dataset--resolution cell and can change the leading method. Global-micro (G) and image-macro (I) Dice yield different winners for \acs{bkai} at $256\times256$ and \acs{isic} at both resolutions, while G and class-macro (C) preserve the winner but reorder up to seven ranks. Full rank matrices are provided in the Appendix.

\begin{table}[ht]
\centering
\caption{Ranking sensitivity to pairwise reaggregation. $\rho$: Spearman correlation; $n_{\Delta}/N$: changed ranks; W: changed winner. 
}
\label{tab:aggregation-sensitivity}
\footnotesize
\setlength{\tabcolsep}{2.8pt}
\begin{adjustbox}{max width=0.47\textwidth}
    \begin{tabular}{llccccccccc}
    \toprule
    \multirow{2}{*}{Dataset} & \multirow{2}{*}{Size} & \multicolumn{3}{c}{G vs. I} & \multicolumn{3}{c}{G vs. C} & \multicolumn{3}{c}{I vs. C} \\
    \cmidrule(lr){3-5}\cmidrule(lr){6-8}\cmidrule(lr){9-11}
     & & $\rho$ & $n_{\Delta}/N$ & W & $\rho$ & $n_{\Delta}/N$ & W & $\rho$ & $n_{\Delta}/N$ & W \\
    \midrule
    \acs{bkai}  & 256 & .95 & 7/11 & Yes & .95 & 7/11 & No  & .95 & 7/11 & Yes \\
    \acs{bkai}  & 512 & .99 & 2/10 & No  & .94 & 7/10 & No  & .95 & 5/10 & No  \\
    \acs{camus}     & 256 & .99 & 2/11 & No  & 1.00 & 0/11 & No  & .99 & 2/11 & No  \\
    \acs{camus}      & 512 & 1.00 & 0/10 & No  & 1.00 & 0/10 & No  & 1.00 & 0/10 & No  \\
    \acs{isic} & 256 & .93 & 5/11 & Yes & 1.00 & 0/11 & No  & .93 & 5/11 & Yes \\
    \acs{isic} & 512 & .93 & 6/10 & Yes & 1.00 & 0/10 & No  & .93 & 6/10 & Yes \\
    \bottomrule
    \end{tabular}
\end{adjustbox}
\end{table}

\begin{figure*}[ht]
  \centering
  \includegraphics[trim={1.7cm 0 0 30},clip, width=\linewidth]{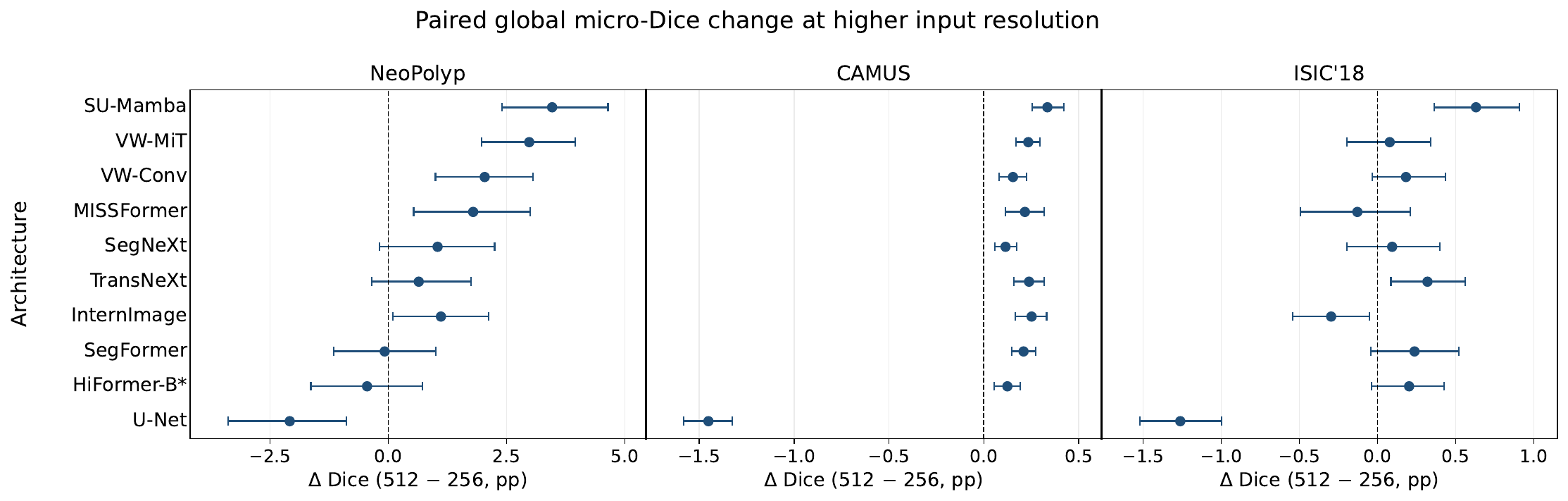}
  \caption{Paired change in global-micro foreground Dice when increasing the input resolution from $256\times256$ to $512\times512$. 
  }
  \label{fig:case_study:input_resolution_sensitivity}
\end{figure*}

\textbf{Higher resolution is not universally beneficial.}
Fig.~\ref{fig:case_study:input_resolution_sensitivity} shows paired global-micro Dice changes from $256\times256$ to $512\times512$. Effects vary by dataset and architecture: \acs{bkai} ranges from a $+$3.46 pp gain for SU-Mamba (95\% CI +2.40 to +4.64) to a $-$2.08 pp loss for U-Net (95\% CI $-$3.38 to $-$0.89); CAMUS shows mostly small gains ($+$0.11 to $+$0.33 pp), while \acs{isic} includes losses for \acs{internimage} ($-$0.30 pp) and \acs{unet} ($-$1.26) pp. 

\textbf{Efficiency profiles differ substantially.}
Table~\ref{tab:inference-efficiency} exposes trade-offs that are invisible from parameter counts alone. At $256\times256$, the models span 9.36--59.89~GMACs and 4.09--74.92~ms per image; at $512\times512$, the available models span 37.43--238.43~GMACs and 13.17--57.46~ms. The table therefore supports selecting an architecture under an explicit deployment constraint, rather than treating DSC as the sole criterion.
More detailed analysis, including a complexity-vs.-DSC-plot, are provided in the Appendix.

\textbf{External transfer is model-/ resolution-dependent.}
Fig.~\ref{fig:case_study:kvasir_ood:XAI} shows same-task transfer from \acs{bkai} to 
Kvasir-SEG. Across 11 models at $256\times256$, strict-majority ensemble DSC ranged from 0.689 (\acs{missformer}) to 0.862 (\acs{transnext}); across the 10 models with larger input size, from 0.687 (\acs{missformer}) to 0.861 (VW-Conv). Fold-level markers indicate checkpoint variation. Resolution effects were architecture-dependent, improving \acs{segformer} and \acs{vwmit} but reducing \acs{unet}, \acs{transnext}, and \acs{swinumamba}. The $256\times256$ case in Fig.~\ref{fig:case_study:kvasir_ood:XAI}, selected by the median architecture-level score, shows consistent recovery of the main polyp region but visible boundary differences.

\begin{figure}[H]
  \centering
  \includegraphics[width=0.45\textwidth]{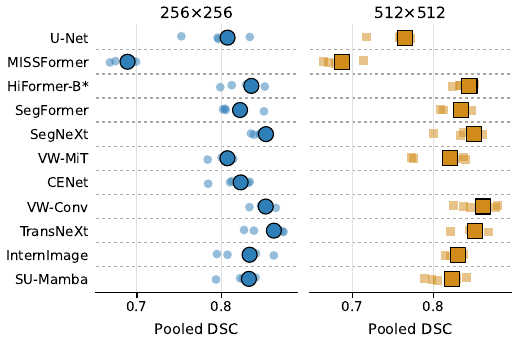}
   \caption{
   Pooled foreground DSC for each evaluated \acs{cv} fold (small, semi-transparent) and the strict-majority ensemble (larger). 
   }
  \label{fig:case_study:kvasir_ood:DSC}
\end{figure}

\begin{table*}[ht]
\centering
\caption{Model-only inference efficiency at batch size one. GMACs denote model-operation counts per forward pass; latency is the median across repeated runs ($\pm$ between-repeat SD); peak memory is the maximum allocated CUDA memory observed across repetitions.}
\label{tab:inference-efficiency}
\small
\begin{threeparttable}
\begin{adjustbox}{max width=\textwidth}
\begin{tabular}{lccccccccc}
\toprule
\multirow{2}{*}{Architecture} & \multirow{2}{*}{Param (M)} & \multicolumn{4}{c}{256$\times$256} & \multicolumn{4}{c}{512$\times$512} \\
\cmidrule(lr){3-6}\cmidrule(lr){7-10}
 & & GMACs & Latency (ms) & TP (img/s) & PM (GiB) & GMACs & Latency (ms) & TP (img/s) & PM (GiB) \\
\midrule
U-Net & 31.03 & 48.18 & 4.09 $\pm$ 0.00 & 244.4 & 0.62 & 192.70 & 13.17 $\pm$ 0.00 & 75.9 & 0.96 \\
MISSFormer & 42.46 & 9.36 & 32.75 $\pm$ 0.13 & 30.5 & 0.70 & 37.43 & 57.46 $\pm$ 0.04 & 17.4 & 1.66 \\
HiFormer-B* & 45.36\tnote{1} & 25.52 & 27.12 $\pm$ 0.18 & 36.9 & 0.72 & 100.03 & 56.55 $\pm$ 0.17 & 17.7 & 0.83 \\
SegFormer & 47.22 & 17.80 & 37.46 $\pm$ 0.18 & 26.7 & 0.86 & 71.21 & 39.27 $\pm$ 0.23 & 25.5 & 1.22 \\
SegNeXt & 48.78 & 16.05 & 32.56 $\pm$ 0.30 & 30.7 & 0.79 & 64.19 & 32.58 $\pm$ 0.10 & 30.7 & 0.85 \\
VW-MiT & 51.42 & 15.55 & 38.79 $\pm$ 0.19 & 25.8 & 0.84 & 62.81 & 41.50 $\pm$ 0.16 & 24.1 & 0.95 \\
CENet & 53.25 & 19.13 & 74.92 $\pm$ 0.40 & 13.3 & 1.41 & -- & -- & -- & -- \\
VW-Conv & 57.01 & 19.21 & 17.52 $\pm$ 0.17 & 57.1 & 0.93 & 77.44 & 24.93 $\pm$ 0.00 & 40.1 & 1.04 \\
TransNeXt & 57.74 & 59.60 & 50.85 $\pm$ 0.14 & 19.7 & 1.02 & 238.43 & 56.88 $\pm$ 0.24 & 17.6 & 1.31 \\
InternImage & 58.37 & 58.57 & 36.17 $\pm$ 0.15 & 27.6 & 1.03 & 234.25 & 41.64 $\pm$ 0.23 & 24.0 & 1.28 \\
SU-Mamba & 59.89 & 39.74 & 21.45 $\pm$ 0.09 & 46.6 & 1.04 & 158.97 & 42.54 $\pm$ 0.01 & 23.5 & 1.34 \\
\bottomrule
\end{tabular}
\end{adjustbox}
\begin{tablenotes}[para,flushleft]
\scriptsize
\item[1] HiFormer-B*: parameter counts differ by input resolution; the reported value is their mean (256$\times$256: 44.62 M; 512$\times$512: 46.09 M).
\end{tablenotes}
\end{threeparttable}
\end{table*}

\section{Discussion}
\label{sec:discussion}

\textbf{From score reporting to inspectable evaluation claims.}
The case study shows that segmentation scores depend on both the estimator and experimental conditions. Reaggregation changed the leading architecture in several dataset--resolution cells, while $512\times512$ input yielded neither consistent gains nor uniform computational costs. Thus, aggregation and resolution choices define the evaluation question rather than a universally optimal setting. \toolnameShort makes these choices explicit and preserves the quantities needed to revisit them after training. 
Inference records further show that parameter count alone does not characterize efficiency: latency and memory depend on the configuration, with \ac{cenet} exceeding available GPU memory at $512\times512$. Together, these capabilities shift benchmarking from isolated scores toward reproducible, inspectable claims under a declared protocol.

\textbf{Threats to validity.}
The current toolbox is scoped to supervised 2D semantic segmentation and is not a general platform for 3D, instance, detection, prompt-based, or deployment workflows. New architectures and datasets still require adapters or conversion to the \texttt{index.csv} 
schema. Configuration hashes and study checkpoints improve traceability, but do not replace complete dataset-version or containerized environment manifests. Likewise, duplicate diagnostics support review but neither prevent leakage automatically nor substitute patient-, study-, or site-level metadata. Post-hoc analyses, including profiling, currently require study-specific maintenance when models, datasets, or resolutions change; the underlying structure is modular, but not yet fully plug-and-play. Fold-wise segmentations can be retained, yet automated boundary-, instance-, calibration-, and probability-based analyses are not yet provided. 
The integrated \ac{hpo} functionality automate executions, not the choice of search space or evaluation protocol. Finally, deterministic execution remains best effort across hardware and software environments.

\textbf{Future work.}
Future work should extend the scope to 3D and instance-level segmentation and build analysis modules that use the already retained masks and optional logits for surface- and object-based metrics and calibrated probabilistic analyses. 
Further priorities are declarative semantic mappings for external evaluation, more complete dataset and execution-environment provenance, and reducing the study-specific maintenance currently required for post-hoc analyses.
Finally, \toolnameShort itself should be evaluated as infrastructure: independent users should add a dataset and architecture from the documentation alone, with onboarding effort and reproduction deviations measured explicitly.

\section{Conclusion}
\label{sec:conclusion}

We introduced \toolname, a configuration-driven PyTorch framework for controlled 2D \ac{mis} benchmarks. It links duplicate-aware data preparation, layered experiment specifications, resumable execution, checkpoint-backed evaluation, and raw-count-first post-hoc analysis. The case study shows that architecture rankings can depend on the aggregation rule and that increasing input resolution yields dataset- and architecture-specific accuracy--efficiency trade-offs. Rather than treating a benchmark result as a standalone score, \toolnameShort preserves the elementary evidence and declared conditions needed to inspect, recompute, and qualify the resulting claim.

{
    \small
    \bibliographystyle{ieeenat_fullname}
    \bibliography{bib-strings,main}

@string(CVPR= {IEEE Conf. Comput. Vis. Pattern Recog.})

@string(ICCV= {Int. Conf. Comput. Vis.})

@string(ECCV= {Eur. Conf. Comput. Vis.})

@string(NIPS= {Adv. Neural Inform. Process. Syst.})

@string(KDD = {Knowledge Discovery \& Data Mining })

@string(CVPR  = {CVPR})

@string(ICCV  = {ICCV})

@string(ECCV  = {ECCV})

@string(NIPS  = {NeurIPS})

@string(TMI = {IEEE Trans. Med. Imaging})

@string(MICCAI = {Med. Image Comput. Comput. Assist. Interv.})

@string(WACV = {IEEE Winter Conf. Appl. Comput. Vis.})

@article{Stupple2019reproducibility,
  title         = {The reproducibility crisis in the age of digital medicine},
  author        = {Aaron Stupple and David Singerman and Leo Anthony Celi},
  year          = {2019},
  journal       = {npj Digit. Medicine},
  volume        = {2},
  doi           = {10.1038/s41746-019-0079-z}
}

@inproceedings{akiba2019optuna,
  title         = {{O}ptuna: A Next-Generation Hyperparameter Optimization Framework},
  author        = {Akiba, Takuya and Sano, Shotaro and Yanase, Toshihiko and Ohta, Takeru and Koyama, Masanori},
  year          = {2019},
  booktitle     = KDD,
  doi           = {10.1145/3292500.3330701}
}

@misc{Borst2026bmvc,
  title         = {{GP-VM\texttimes{}SMA}: Benchmarking General-Purpose Vision Models and Specialized Architectures for {2D} Medical Image Segmentation},
  author        = {Anonymous},
  year          = {2026},
  note          = {Accepted at British Machine Vision Conference 2026, Submission ID 482. Supplied as supplemental material {\tt BMVC2026-Submission-482.pdf}}
}

@article{antonelli2022msd,
  title         = {The Medical Segmentation Decathlon},
  author        = {Antonelli, Michela and Reinke, Annika and Bakas, Spyridon and others},
  year          = {2022},
  journal       = {Nature Communications},
  volume        = {13},
  pages         = {4128},
  doi           = {10.1038/s41467-022-30695-9}
}

@article{baker1500ScientistsLift2016,
  title         = {1,500 Scientists Lift the Lid on Reproducibility},
  author        = {Baker, Monya},
  year          = {2016},
  journal       = {Nature},
  publisher     = {Nature Publishing Group},
  volume        = {533},
  number        = {7604},
  pages         = {452--454},
  doi           = {10.1038/533452a}
}

@inproceedings{bozorgpour2025cenet,
  title         = {{CENet}: Context Enhancement Network for Medical Image Segmentation},
  author        = {Bozorgpour, Afshin and Kolahi, Sina Ghorbani and Azad, Reza and Hacihaliloglu, Ilker and Merhof, Dorit},
  year          = {2025},
  booktitle     = MICCAI,
  doi           = {10.1007/978-3-032-04927-8_12}
}

@inproceedings{hu2023SwinUnet,
  title         = {Swin-Unet: Unet-Like Pure Transformer for~Medical Image Segmentation},
  author        = {Cao, Hu and Wang, Yueyue and Chen, Joy and Jiang, Dongsheng and Zhang, Xiaopeng and others},
  year          = {2023},
  booktitle     = ECCV,
  doi           = {10.1007/978-3-031-25066-8_9}
}

@misc{cardoso2022monai,
  title         = {{MONAI}: An Open-Source Framework for Deep Learning in Healthcare},
  author        = {Cardoso, M. Jorge and Li, Wenqi and Brown, Richard and others},
  year          = {2022},
  doi           = {10.48550/arXiv.2211.02701}
}

@inproceedings{christodoulou2024confidence,
  title         = {Confidence Intervals Uncovered: Are We Ready for Real-World Medical Imaging {AI}?},
  author        = {Christodoulou, Evangelia and Reinke, Annika and Houhou, Rola and others},
  year          = {2024},
  booktitle     = MICCAI,
  doi           = {10.1007/978-3-031-72117-5_12}
}

@inproceedings{codella2019isic,
  title         = {Skin lesion analysis toward melanoma detection: A challenge at the 2017 International symposium on biomedical imaging (ISBI), hosted by the international skin imaging collaboration (ISIC)},
  author        = {Codella, Noel C. F. and Gutman, David and Celebi, M. Emre and Helba, Brian and Marchetti, Michael A. and others},
  year          = {2018},
  booktitle     = {2018 IEEE 15th International Symposium on Biomedical Imaging (ISBI 2018)},
  doi           = {10.1109/isbi.2018.8363547}
}

@inproceedings{Cohen2022xrv,
  title         = {TorchXRayVision: A library of chest X-ray datasets and models},
  author        = {Cohen, Joseph Paul and Viviano, Joseph D. and Bertin, Paul and Morrison, Paul and Torabian, Parsa and others},
  year          = {2022},
  booktitle     = {Medical Imaging with Deep Learning}
}

@inproceedings{guo2022segnext,
  title         = {{SegNeXt}: Rethinking Convolutional Attention Design for Semantic Segmentation},
  author        = {Guo, Meng-Hao and Lu, Cheng-Ze and Hou, Qibin and Liu, Zheng-Ning and Cheng, Ming-Ming and others},
  year          = {2022},
  booktitle     = NIPS,
  doi           = {10.5555/3600270.3600354}
}

@inproceedings{heidari2023hiformer,
  title         = {{HiFormer}: Hierarchical Multi-Scale Representations Using Transformers for Medical Image Segmentation},
  author        = {Heidari, Moein and Kazerouni, Amirhossein and Soltany, Milad and Azad, Reza and Aghdam, Ehsan Khodapanah and others},
  year          = {2023},
  booktitle     = WACV
}

@article{huang2022missformer,
  title         = {{MISSFormer}: An Effective Transformer for {2D} Medical Image Segmentation},
  author        = {Huang, Xiaohong and Deng, Zhifang and Li, Dandan and Yuan, Xueguang and Fu, Ying},
  year          = {2023},
  journal       = TMI,
  volume        = {42},
  number        = {5},
  pages         = {1484--1494},
  doi           = {10.1109/tmi.2022.3230943}
}

@inproceedings{hutchinsonEvaluationGapsMachine2022,
  title         = {Evaluation {{Gaps}} in {{Machine Learning Practice}}},
  author        = {Hutchinson, Ben and Rostamzadeh, Negar and Greer, Christina and Heller, Katherine and Prabhakaran, Vinodkumar},
  year          = {2022},
  booktitle     = {{{ACM Conference}} on {{Fairness Accountability}} and {{Transparency}}},
  doi           = {10.1145/3531146.3533233}
}

@article{isensee2021nnunet,
  title         = {{nnU-Net}: A Self-Configuring Method for Deep Learning-Based Biomedical Image Segmentation},
  author        = {Isensee, Fabian and Jaeger, Paul F. and Kohl, Simon A. A. and Petersen, Jens and Maier-Hein, Klaus H.},
  year          = {2021},
  journal       = {Nature Methods},
  volume        = {18},
  pages         = {203--211},
  doi           = {10.1038/s41592-020-01008-z}
}

@article{chen2024TransUNet,
  title         = {TransUNet: Rethinking the U-Net architecture design for medical image segmentation through the lens of transformers},
  author        = {Jieneng Chen and Jieru Mei and Xianhang Li and Yongyi Lu and Qihang Yu and others},
  year          = {2024},
  journal       = {Medical Image Anal.},
  volume        = {97},
  pages         = {103280},
  doi           = {10.1016/j.media.2024.103280}
}

@article{kapoor2023leakage,
  title         = {Leakage and the Reproducibility Crisis in Machine-Learning-Based Science},
  author        = {Kapoor, Sayash and Narayanan, Arvind},
  year          = {2023},
  journal       = {Patterns},
  volume        = {4},
  number        = {9},
  pages         = {100804},
  doi           = {10.1016/j.patter.2023.100804}
}

@article{kus2024medsegbench,
  title         = {{MedSegBench}: A Comprehensive Benchmark for Medical Image Segmentation in Diverse Data Modalities},
  author        = {Ku{\c{s}}, Zeki and Aydin, Musa},
  year          = {2024},
  journal       = {Scientific Data},
  volume        = {11},
  doi           = {10.1038/s41597-024-04159-2}
}

@article{leclerc2019camus,
  title         = {Deep Learning for Segmentation Using an Open Large-Scale Dataset in {2D} Echocardiography},
  author        = {Leclerc, Sarah and Smistad, Erik and Pedrosa, Jo{\~a}o and others},
  year          = {2019},
  journal       = TMI,
  volume        = {38},
  pages         = {2198--2210}
}

@misc{liaw2018tune,
  title         = {Tune: A Research Platform for Distributed Model Selection and Training},
  author        = {Liaw, Richard and Liang, Eric and Nishihara, Robert and Moritz, Philipp and Gonzalez, Joseph E and others},
  year          = {2018},
  doi           = {10.48550/arXiv.1807.05118}
}

@inproceedings{liu2024swinumamba,
  title         = {{Swin-UMamba}: Mamba-Based {UNet} with {ImageNet}-Based Pretraining},
  author        = {Liu, Jiarun and Yang, Hao and Zhou, Hong-Yu and others},
  year          = {2024},
  booktitle     = MICCAI,
  doi           = {10.1007/978-3-031-72114-4_59}
}

@article{ma2024SegmentAnythingMedical,
  title         = {Segment Anything in Medical Images},
  author        = {Ma, Jun and He, Yuting and Li, Feifei and Han, Lin and You, Chenyu and others},
  year          = {2024},
  journal       = {Nature Communications},
  volume        = {15},
  number        = {1},
  pages         = {654},
  doi           = {10.1038/s41467-024-44824-z}
}

@article{maierhein2018rankings,
  title         = {Why Rankings of Biomedical Image Analysis Competitions Should Be Interpreted with Care},
  author        = {Maier-Hein, Lena and Eisenmann, Matthias and Reinke, Annika and Onogur, Sinan and Stankovic, Marko and others},
  year          = {2018},
  journal       = {Nature Communications},
  volume        = {9},
  pages         = {5217},
  doi           = {10.1038/s41467-018-07619-7}
}

@misc{mmseg2020,
  title         = {{MMSegmentation}: OpenMMLab Semantic Segmentation Toolbox and Benchmark},
  author        = {MMSegmentation Contributors},
  year          = {2020},
  howpublished  = {\url{https://github.com/open-mmlab/mmsegmentation}}
}

@inproceedings{bkaiigh2021,
  title         = {NeoUNet : Towards Accurate Colon Polyp Segmentation and~Neoplasm Detection},
  author        = {Ngoc Lan, Phan and An, Nguyen Sy and Hang, Dao Viet and Long, Dao Van and Trung, Tran Quang and others},
  year          = {2021},
  booktitle     = {Advances in Visual Computing},
  doi           = {10.1007/978-3-030-90436-4_2}
}

@article{perezgarcia2021torchio,
  title         = {{TorchIO}: A Python Library for Efficient Loading, Preprocessing, Augmentation and Patch-Based Sampling of Medical Images in Deep Learning},
  author        = {P{\'e}rez-Garc{\'i}a, Fernando and Sparks, Rachel and Ourselin, S{\'e}bastien},
  year          = {2021},
  journal       = {Computer Methods and Programs in Biomedicine},
  volume        = {208},
  pages         = {106236},
  doi           = {10.1016/j.cmpb.2021.106236}
}

@article{fets_tool,
  title         = {The federated tumor segmentation (FeTS) tool: an open-source solution to further solid tumor research},
  author        = {Pati, Sarthak and Baid, Ujjwal and Edwards, Brandon and Sheller, Micah J and Foley, Patrick and others},
  year          = {2022},
  journal       = {Physics in Medicine \& Biology},
  doi           = {10.1088/1361-6560/ac9449}
}

@misc{Iakubovskii:2019,
  title         = {Segmentation Models Pytorch},
  author        = {Pavel Iakubovskii},
  year          = {2019},
  journal       = {GitHub repository},
  publisher     = {GitHub},
  howpublished  = {\url{https://github.com/qubvel/segmentation\_models.pytorch}}
}

@article{pineau2021improving,
  title         = {Improving Reproducibility in Machine Learning Research: A Report from the {NeurIPS} 2019 Reproducibility Program},
  author        = {Pineau, Joelle and Vincent-Lamarre, Philippe and Sinha, Koustuv and Larivi\`ere, Vincent and Beygelzimer, Alina and others},
  year          = {2021},
  journal       = {Journal of Machine Learning Research},
  volume        = {22},
  number        = {1},
  pages         = {1--20},
  doi           = {10.5555/3546258.3546422}
}

@inproceedings{mostafijur2023CascadedAttentionDecoding,
  title         = {Medical Image Segmentation via Cascaded Attention Decoding},
  author        = {Rahman, Md Mostafijur and Marculescu, Radu},
  year          = {2023},
  booktitle     = WACV,
  doi           = {10.1109/wacv56688.2023.00616}
}

@inproceedings{mostafijur2024EMCAD,
  title         = {EMCAD: Efficient Multi-Scale Convolutional Attention Decoding for Medical Image Segmentation},
  author        = {Rahman, Md Mostafijur and Munir, Mustafa and Marculescu, Radu},
  year          = {2024},
  booktitle     = CVPR,
  doi           = {10.1109/cvpr52733.2024.01118}
}

@inproceedings{jha2019kvasir,
  title         = {Kvasir-{SEG}: A segmented polyp dataset},
  author        = {Jha, Debesh and Smedsrud, Pia H and Riegler, Michael A and Halvorsen, P{\aa}l and others},
  booktitle     = {MMM},
  year          = {2019},
  doi           = {10.1007/978-3-030-37734-2_37}
}

@article{reinke2024metrics,
  title         = {Metrics Reloaded: Recommendations for Image Analysis Validation},
  author        = {Reinke, Annika and Tizabi, Minu D. and Sudre, Carole H. and Eisenmann, Matthias and others},
  year          = {2024},
  journal       = {Nature Methods},
  volume        = {21},
  pages         = {48--53},
  doi           = {10.1038/s41592-023-02151-z}
}

@article{renard2020variability,
  title         = {Variability and Reproducibility in Deep Learning for Medical Image Segmentation},
  author        = {Renard, F{\'e}lix and Guedria, Soulaimane and De Palma, Noel and Vuillerme, Nicolas},
  year          = {2020},
  journal       = {Scientific Reports},
  volume        = {10},
  doi           = {10.1038/s41598-020-69920-0}
}

@inproceedings{ronneberger2015unet,
  title         = {{U-Net}: Convolutional Networks for Biomedical Image Segmentation},
  author        = {Ronneberger, Olaf and Fischer, Philipp and Brox, Thomas},
  year          = {2015},
  booktitle     = MICCAI,
  doi           = {10.1007/978-3-319-24574-4_28}
}

@article{jiacheng2025VM-UNet,
  title         = {VM-UNet: Vision Mamba UNet for Medical Image Segmentation},
  author        = {Ruan, Jiacheng and Li, Jincheng and Xiang, Suncheng},
  year          = {2025},
  journal       = {ACM Trans. Multimedia Comput. Commun. Appl.},
  doi           = {10.1145/3767748}
}

@inproceedings{gradcam,
  title         = {Grad-CAM: Visual Explanations from Deep Networks via Gradient-Based Localization},
  author        = {Selvaraju, Ramprasaath R. and Cogswell, Michael and Das, Abhishek and Vedantam, Ramakrishna and Parikh, Devi and others},
  year          = {2017},
  booktitle     = ICCV,
  doi           = {10.1109/iccv.2017.74}
}

@article{semmelrock2025reproducibility,
  title         = {Reproducibility in machine-learning-based research: Overview, barriers, and drivers},
  author        = {Semmelrock, Harald and Ross-Hellauer, Tony and Kopeinik, Simone and Theiler, Dieter and Haberl, Armin and others},
  year          = {2025},
  journal       = {AI Magazine},
  volume        = {46},
  number        = {2},
  pages         = {e70002},
  doi           = {https://doi.org/10.1002/aaai.70002}
}

@inproceedings{shi2024transnext,
  title         = {{TransNeXt}: Robust Foveal Visual Perception for Vision Transformers},
  author        = {Shi, Dai},
  year          = {2024},
  booktitle     = CVPR,
  doi           = {10.1109/cvpr52733.2024.01683}
}

@inproceedings{tang2022SwinTransformers,
  title         = {Self-Supervised Pre-Training of Swin Transformers for 3D Medical Image Analysis},
  author        = {Tang, Yucheng and Yang, Dong and Li, Wenqi and Roth, Holger R. and Landman, Bennett and others},
  year          = {2022},
  booktitle     = CVPR,
  doi           = {10.1109/cvpr52688.2022.02007}
}

@inproceedings{liaoAreWeLearning2021,
  title         = {Are We Learning Yet? {A} Meta Review of Evaluation Failures Across Machine Learning},
  author        = {Thomas Liao and Rohan Taori and Inioluwa Deborah Raji and Ludwig Schmidt},
  year          = {2021},
  booktitle     = NIPS
}

@inproceedings{wolf-etal-2020-transformers,
  title         = {Transformers: State-of-the-Art Natural Language Processing},
  author        = {Thomas Wolf and Lysandre Debut and Victor Sanh and Julien Chaumond and Clement Delangue and others},
  year          = {2020},
  booktitle     = {Conference on Empirical Methods in Natural Language Processing},
  doi           = {10.18653/v1/2020.emnlp-demos.6}
}

@article{tschandl2018ham10000,
  title         = {The {HAM10000} Dataset, a Large Collection of Multi-Source Dermatoscopic Images of Common Pigmented Skin Lesions},
  author        = {Tschandl, Philipp and Rosendahl, Cliff and Kittler, Harald},
  year          = {2018},
  journal       = {Scientific Data},
  volume        = {5},
  doi           = {10.1038/sdata.2018.161}
}

@article{varoquaux2022machine,
  title         = {Machine Learning for Medical Imaging: Methodological Failures and Recommendations for the Future},
  author        = {Varoquaux, Ga{\"e}l and Cheplygina, Veronika},
  year          = {2022},
  journal       = {npj Digital Medicine},
  volume        = {5},
  doi           = {10.1038/s41746-022-00592-y}
}

@inproceedings{wang2023internimage,
  title         = {{InternImage}: Exploring Large-Scale Vision Foundation Models with Deformable Convolutions},
  author        = {Wang, Wenhai and Dai, Jifeng and Chen, Zhe and others},
  year          = {2023},
  booktitle     = CVPR,
  doi           = {10.1109/cvpr52729.2023.01385}
}

@article{Wasserthal_2023,
  title         = {TotalSegmentator: Robust Segmentation of 104 Anatomic Structures in CT Images},
  author        = {Wasserthal, Jakob and Breit, Hanns-Christian and Meyer, Manfred T. and Pradella, Maurice and Hinck, Daniel and others},
  year          = {2023},
  journal       = {Radiology: Artificial Intelligence},
  volume        = {5},
  number        = {5},
  doi           = {10.1148/ryai.230024}
}

@inproceedings{xiao2018upernet,
  title         = {Unified Perceptual Parsing for Scene Understanding},
  author        = {Xiao, Tete and Liu, Yingcheng and Zhou, Bolei and Jiang, Yuning and Sun, Jian},
  year          = {2018},
  booktitle     = ECCV,
  doi           = {10.1007/978-3-030-01228-1_26}
}

@inproceedings{xie2021segformer,
  title         = {{SegFormer}: Simple and Efficient Design for Semantic Segmentation with Transformers},
  author        = {Xie, Enze and Wang, Wenhai and Yu, Zhiding and Anandkumar, Anima and Alvarez, Jose M. and others},
  year          = {2021},
  booktitle     = NIPS,
  doi           = {10.5555/3540261.3541185}
}

@misc{liu2021paddleseg,
  title         = {PaddleSeg: A High-Efficient Development Toolkit for Image Segmentation},
  author        = {Yi Liu and Lutao Chu and Guowei Chen and Zewu Wu and Zeyu Chen and others},
  year          = {2021},
  doi           = {10.48550/arXiv.2101.06175}
}

@article{zenk_towards_2025,
  title         = {Towards fair decentralized benchmarking of healthcare {AI} algorithms with the {Federated} {Tumor} {Segmentation} ({FeTS}) challenge},
  author        = {Zenk, Maximilian and Baid, Ujjwal and Pati, Sarthak and Linardos, Akis and Edwards, Brandon and others},
  year          = {2025},
  journal       = {Nature Communications},
  volume        = {16},
  number        = {1},
  pages         = {6274},
  doi           = {10.1038/s41467-025-60466-1}
}

@article{zhou2021ModelsGenesis,
  title         = {Models Genesis},
  author        = {Zongwei Zhou and Vatsal Sodha and Jiaxuan Pang and Michael B. Gotway and Jianming Liang},
  year          = {2021},
  journal       = {Medical Image Analysis},
  volume        = {67},
  pages         = {101840},
  doi           = {https://doi.org/10.1016/j.media.2020.101840}
}
}

\end{document}